%% file: main.tex
\documentclass[10pt,twocolumn,letterpaper]{article}

\usepackage{cvpr}

\definecolor{cvprblue}{rgb}{0.21,0.49,0.74}
\usepackage[pagebackref,breaklinks,colorlinks,allcolors=cvprblue]{hyperref}
\usepackage{amsmath}
\usepackage{xcolor}
\usepackage{multirow}
\usepackage{booktabs}
\usepackage{tabularx} 
\usepackage{graphicx}
\usepackage{graphicx}
\usepackage[export]{adjustbox} 
\usepackage{graphicx} 
\usepackage{caption}
\usepackage{adjustbox}
\usepackage{fix-cm}
\definecolor{darkgray}{rgb}{0.3, 0.3, 0.3}
\definecolor{darkred}{RGB}{197,53,48}
\definecolor{gray666}{RGB}{102,102,102}


\def\confName{CVPR}
\def\confYear{2025}

\title{
    \includegraphics[height=1.35em,valign=c]{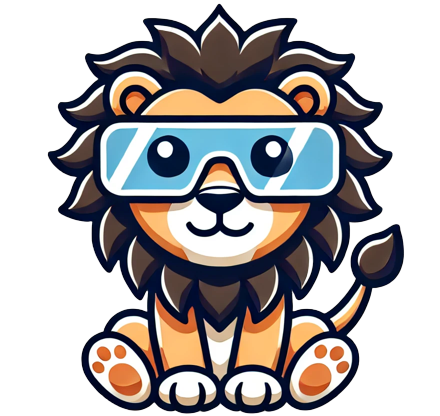}~LION-FS: Fast \& Slow Video-Language Thinker as Online Video Assistant
}

\author{
    Wei Li, Bing Hu, Rui Shao\thanks{Corresponding author.}, Leyang Shen, Liqiang Nie\\
    Harbin Institute of Technology, Shenzhen \\
    {\tt\small liwei2024@stu.hit.edu.cn \quad shaorui@hit.edu.cn}\\
\texttt{\normalsize{\url{https://github.com/JiuTian-VL/LION-FS}}}
}

\begin{document}
\maketitle

\input{sec/0_abstract}

\input{sec/1_intro}

\input{sec/2_related_works}
\input{sec/3_method}

\input{sec/4_experience}

\input{sec/5_conclusion}
\input{sec/X_suppl}

\clearpage
{
    \small
    \bibliographystyle{ieeenat_fullname} 
    \bibliography{main}
}

\end{document}

%% file: sec/0_abstract.tex
\begin{abstract}

\vspace{-5pt}
First-person video assistants are highly anticipated to enhance our daily lives through online video dialogue. However, existing online video assistants often sacrifice assistant efficacy for real-time efficiency by processing low-frame-rate videos with coarse-grained visual features. To overcome the trade-off between efficacy and efficiency, we propose ``\textbf{F}ast \& \textbf{S}low Video-Language Thinker" as on\textbf{LI}ne vide\textbf{O} assista\textbf{N}t, \textbf{LION-FS}, achieving real-time, proactive, temporally accurate, and contextually precise responses. LION-FS adopts a two-stage optimization strategy: \textbf{1) Fast Path: Routing-Based Response Determination} evaluates frame-by-frame whether an immediate response is necessary. To enhance response determination accuracy and handle higher frame-rate inputs efficiently, we employ Token Aggregation Routing to dynamically fuse spatiotemporal features without increasing token numbers, while utilizing Token Dropping Routing to eliminate redundant features, and \textbf{2) Slow Path: Multi-granularity Keyframe Augmentation} optimizes keyframes during response generation. To provide comprehensive and detailed responses beyond atomic actions constrained by training data, fine-grained spatial features and human-environment interaction features are extracted through multi-granular pooling. They are further integrated into a meticulously designed multimodal Thinking Template to guide more precise response generation. Comprehensive evaluations of online video tasks demonstrate that LION-FS achieves state-of-the-art efficacy and efficiency. 

\end{abstract}

%% file: sec/1_intro.tex
\begin{figure}[t]
  \centering
   \includegraphics[width=1.0\linewidth]{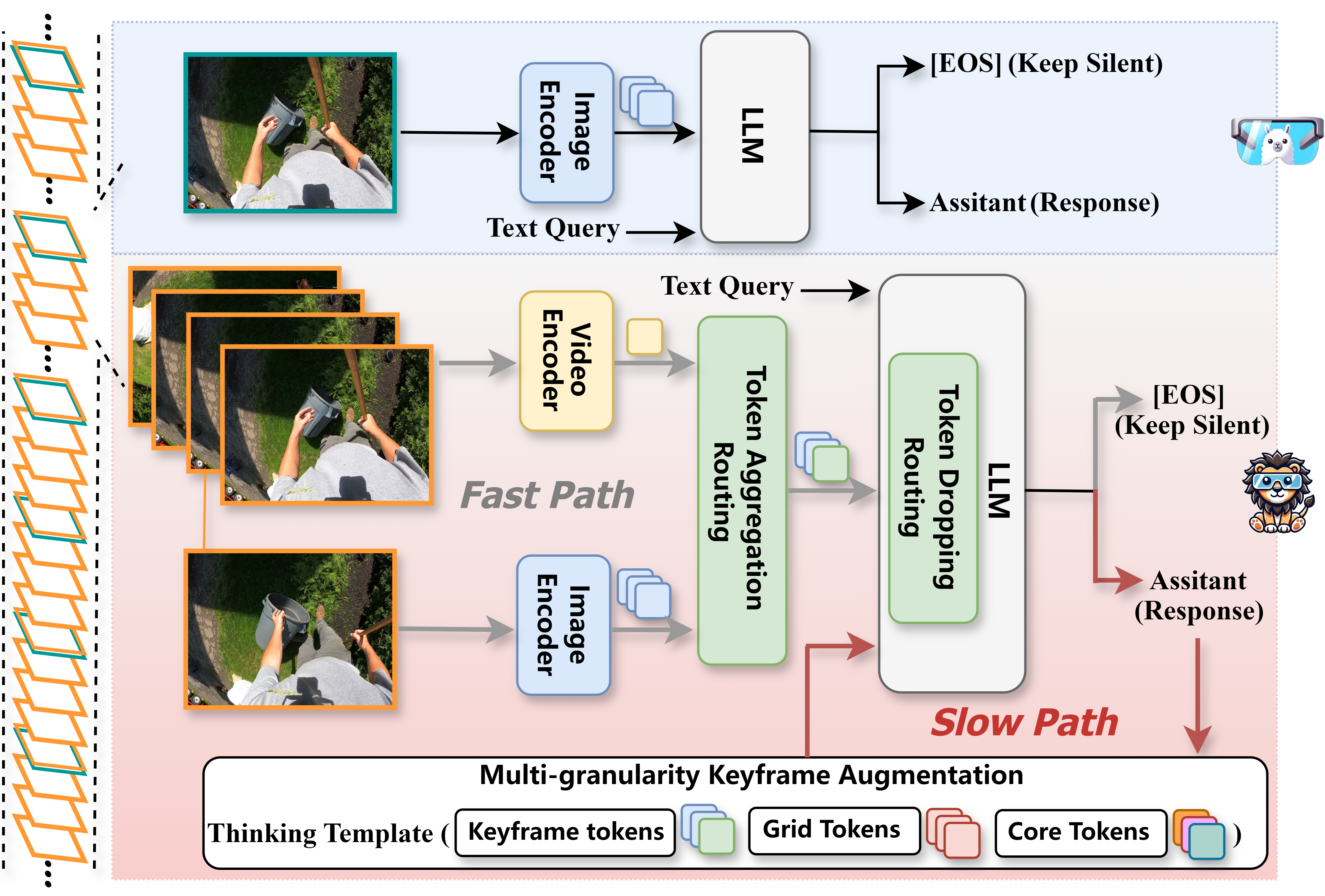}
   \vspace{-2em}
   \caption{Comparison between LIVE~\cite{VideoLLM-online}\includegraphics[height=0.9em]{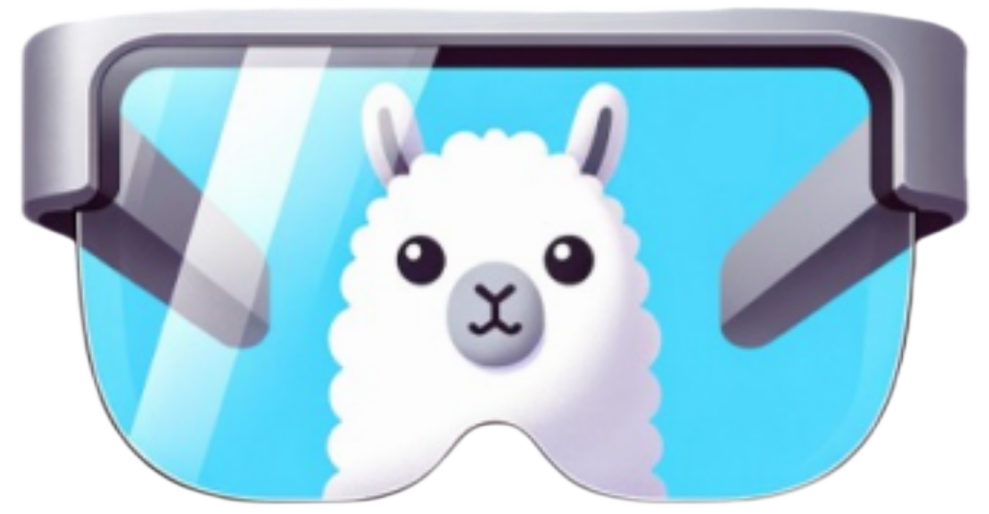} and LION-FS\includegraphics[height=1.5em]{fig/LION-FS.png}. LIVE processes low-frame-rate videos using coarse-grained image tokens, resulting in suboptimal accuracy in response. LION-FS, by efficiently handling high-frame-rate videos through Fast-Path dynamical spatiotemporal fusion and Slow-Path multi-granular keyframe augmentation, significantly enhances response determination accuracy and content precision.}
   \vspace{-10pt}
   \label{fig:fig1}
\end{figure}

% \begin{figure*}[t]
%   \centering
%    \includegraphics[width=1.0\linewidth]{fig/demo.png}
%    \caption{Comparison of LIVE~\cite{VideoLLM-online}\includegraphics[height=0.9em]{fig/videollmonline.png} and LION-FS\includegraphics[height=1.5em]{fig/LION-FS.png} visualizations on the Ego4D and Ego-Exo4D datasets. The purple highlights indicate imprecise responses, while the red highlights denote incorrect responses or errors in response determinantion. LION-FS demonstrates more accurate response determination and delivers responses with more precise content and greater temporal coherence.}
%    \label{fig:demo}
% \end{figure*}

\vspace{-15pt}
\section{Introduction}
\label{sec:intro}

\vspace{-5pt}
In current popular smart glasses~\cite{Google-Glass, Meta-Smart-Glasses, Vuzix-Blade} or head-mounted devices~\cite{HoloLens-2, RealWear}, although most integrate AI applications such as voice assistant~\cite{Siri-apple, Amazon-alexa, Salmonn} and gesture recognition~\cite{Enabling, GazePointAR}, online video assistant~\cite{VideoLLM-online, videollm-mod} have yet to see mature implementation. A primary challenge is that online video assistants require continuous reception of first-person perspective video streams. Additionally, they must have the ability to handle user queries in real-time and proactively provide professional responses or guidance. This imposes high demands on both efficacy and efficiency.

Although existing video understanding works~\cite{Video-llama, Videochat, Moviechat, Video-chatgpt, Llava-onevision, Ma-lmm} achieve high performance in offline scenarios for tasks like video question answering~\cite{vvqa, vvqa1, Flash-VStream, fei2024video}, captioning~\cite{caption1, caption2, caption3}, and spatiotemporal localization~\cite{Ground1, Ground2, Ground3, MMNeedle}, they are unsuitable for the online paradigm of video assistant. VideoLLM-online \cite{VideoLLM-online} is a pioneering work that introduces the video streaming dialogue framework LIVE to video assistant, as shown in Figure~\ref{fig:fig1}. LIVE continuously receives incoming video streams, autonomously determines response timing based on user queries, and provides concise responses. Despite its innovation, LIVE has significant limitations: \textit{\textbf{1) Limited accuracy in response determination.}} LIVE's visual encoding is restricted to low frame-rate image features, which hinders the ability of Multimodal Large Language Models (MLLMs) to learn and capture inter-frame temporal relationships effectively. \textit{\textbf{2) Lack of precision in responses.}} By retaining a fixed and limited number of tokens for all video frames without leveraging the unique characteristics of the first-person perspective, LIVE fails to capture adaptive and detailed egocentric visual information. This inadequate extraction of visual information leads to suboptimal video-language fusion and thus generates imprecise responses. 
\textit{\textbf{3) Inefficiency in training and inference.}} 
% LIVE expands tokens for all frames, thus trading off efficiency for performance. However, redundant visual tokens should be discarded during the response determination phase without expansion. Consequently, token expansion is necessary only for keyframes during response generation. 
LIVE expands tokens for all frames to enhance efficacy, but this minimal expansion proves insufficient. Token expansion is not required during the relatively simple response determination phase; instead, substantial token expansion is necessary only for keyframes during response generation.

To address these challenges, as shown in Figure~\ref{fig:fig1}, we propose ``\textbf{F}ast \& \textbf{S}low Video-Language Thinker" as on\textbf{LI}ne vide\textbf{O} assista\textbf{N}t, \textbf{LION-FS}, which integrates a fast-slow reasoning approach to simulate human thinking and response processes. 
% It browses video content in real-time, autonomously determines when to respond, and provides accurate, coherent answers. 
LION-FS uses a Fast \& Slow two-path optimization scheme, combining fine-tuning with training-free methods, to significantly improve both efficacy and efficiency.
\textit{\textbf{1) Fast Path process for Routing-Based Response Determination}}. To improve the accuracy of response determination, we incorporate not only the extensive visual knowledge from general-purpose encoders but also two additional information components: (i) denser temporal features and (ii) first-person perspective features. To this end, we design a Token Aggregation Routing module. It adaptively aggregates video features extracted by a first-person video encoder from high-frame-rate videos. It also combines these with image features extracted by a general-purpose third-person encoder from low-frame-rate videos. This approach effectively consolidates the advantages of these distinct feature types without increasing the token numbers. 
To further enhance the efficiency of determination, a Token Dropping Routing module is introduced to adaptively discard redundant tokens, thereby sparsifying the decoding computations within the LLM.
\textit{\textbf{2) Slow Path for Multi-Granularity Keyframe Augmentation}}. To enhance response precision, we define the current frame where a response is determined as keyframe and apply multi-granularity augmentation on it: (i) Global uniform augmentation for Grid Tokens: the keyframe is divided into multiple grids, which are pooled at the same granularity as original frames and sequentially concatenated to form Grid Tokens (fine-grained spatial features); (ii) Local adaptive augmentation for Box Tokens: we detect hands and interacting objects within the keyframe, and select tokens within bounding boxes from the patch tokens, followed by pooling to obtain Box Tokens (local features of action occurrence). Finally, these augmented tokens are integrated into a designed Thinking Template, serving as a multimodal prompt to guide more precise and fine-grained response generation.
% To enhance the assessment of response occurrence and improve the precise of responses, we introduce the \textit{\textbf{1) Fast-Path process for Routing-Based Response Determination}}. We employ a third-person image encoder to extract rich spatial features, leveraging extensive visual knowledge and robust generalization capabilities. Simultaneously, a first-person video encoder is utilized to capture first-person features and aggregate adjacent frame characteristics, thereby mitigating the temporal information loss caused by low frame rates. The dual-encoded features are transmitted sequentially to Token Aggregation Routing and Token Dropping Routing, performing weighted aggregation based on user query types to effectively balance the strengths of different features, and adaptively selecting and discarding redundant spatial tokens. To optimize the assistant’s responses and ensure they go beyond the brief, atomic action descriptions typical of the training data, we introduce the \textit{\textbf{2) Slow-Path process for Multi-Granularity Keyframe Augmentation}}. This process enables the assistant to provide coherent, contextually rich answers, enhanced with spatiotemporal granularity. We augment keyframe tokens into Grid Tokens (fine-grained spatial features) and Core Tokens (local features of action occurrence), integrating them into a meticulously designed Thinking Template that serves as a multimodal prompt to guide response generation in online video dialogue. 

We summarize our contributions as follows:
\begin{itemize}
    \item We propose LION-FS, an innovative online video assistant that mimics human cognitive processes by employing fast thinking for simple response determination and slow thinking for complex response generation.
    \item  We develop a Fast \& Slow framework combining fine-tuning with training-free methods. Fast Path improves response determination via routing-based token aggregation and dropping; Slow Path enhances response precision using multi-granularity keyframe augmentation. 
    \item Comprehensive evaluations on the Ego4D and Ego-Exo4D datasets reveal that LION-FS achieves optimal performance and efficiency, outperforming existing methods in online first-person video dialogue tasks.
\end{itemize}

%% file: sec/2_related_works.tex
\begin{figure*}[t]
  \centering
   \includegraphics[width=1.0\linewidth]{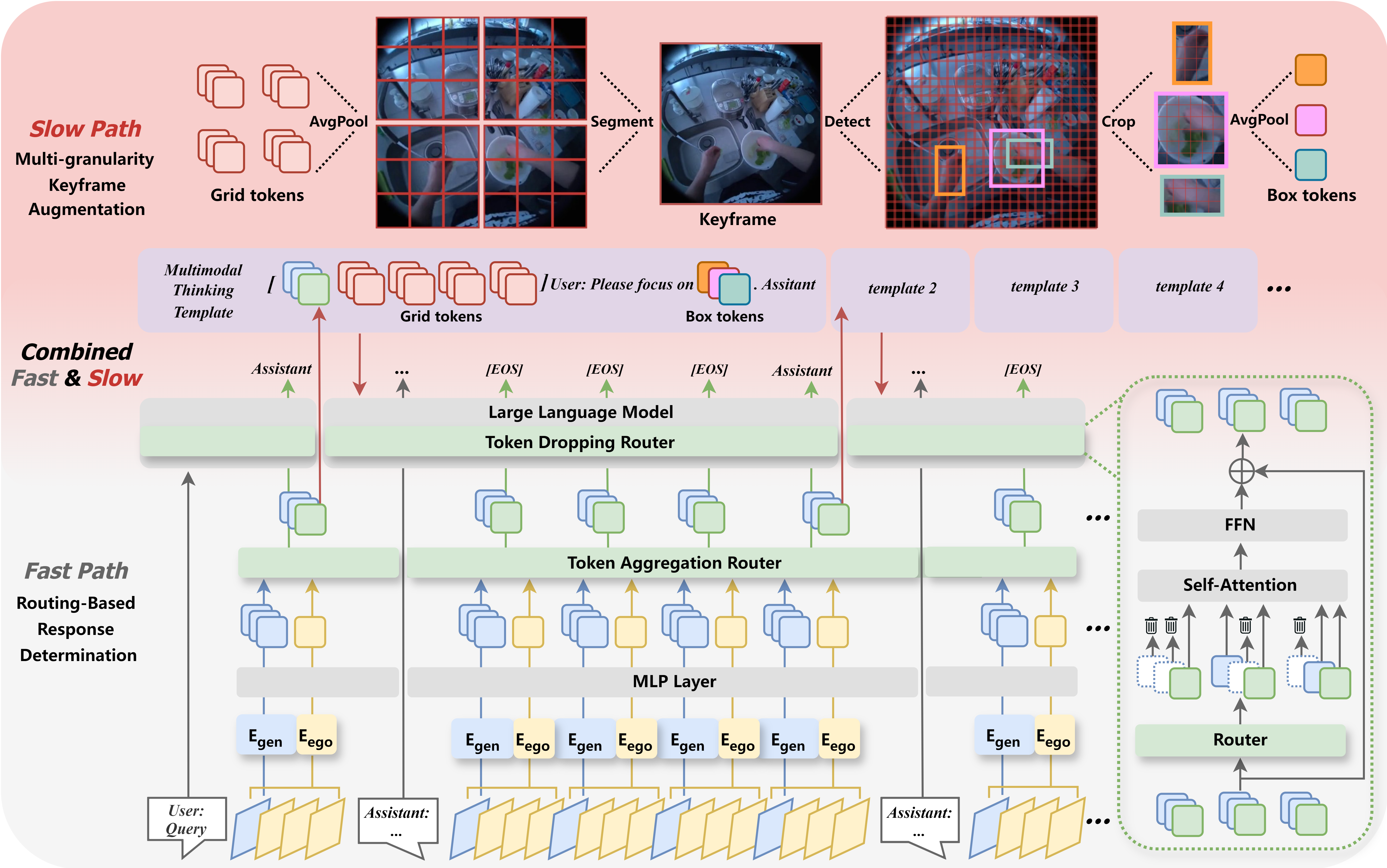}
   \caption{The whole framework of LION-FS. \textcolor{darkgray}{Fast Path} enables high-frame-rate video stream reception, allowing real-time determination of whether a response is required. $E_{gen}$ (SigLIP~\cite{siglip}) extracts general spatial features from 2 FPS frames, while $E_{ego}$ (EgoVLPv2~\cite{egovlpv2}) captures first-person temporal features from 8 FPS frames. These are temporally aligned, weighted through the Token Aggregation Router, and then filtered for redundancy by the Token Dropping Router. \textcolor{darkred}{Slow Path} enhances keyframes with rich information, performing multi-granularity augmentation that includes fine-grained global tokens (Grid Tokens) and action-related local tokens (Box Tokens), which are injected into the Multimodal Thinking Template to guide the assistant in generating more precise responses.}
   \vspace{-10pt}
   \label{fig:framework}
\end{figure*}

\vspace{-5pt}
\section{Related Works}

% \textbf{Online Visual Dialogue.} Most visual dialogue research~\cite{Large-scale, Vd-bert, Video-llama, Videochat, Moviechat, Video-chatgpt, Deepspeed-visualchat} inputs the complete image or video into Multimodal Large Language Models (MLLMs)~\cite{blip2, llava, minigpt4, mplug-owl2, mplug-owl3, blip-instruction} in a single instance for comprehension, thereby supporting offline multi-turn dialogues based exclusively on text. Although some multimodal approaches~\cite{Visual-chatgpt, Llava-next-interleave} have extended dialogue to incorporate interleaved text and images, they have not addressed the challenges associated with video dialogue. These offline methods require complete visual inputs and clearly defined response timings, which do not meet the demands of real-time online video assistants. Online captioning~\cite{Streaming} and video understanding~\cite{Flash-VStream, Streaming-long} can process continuous streams but struggle with the autonomous, real-time responses essential for video assistance. Videollm-online\cite{VideoLLM-online} pioneers a new paradigm for online video stream dialogue, establishing a foundational framework for video assistants. Despite challenges such as inaccurate response determination, overly brief answers reliant on current frame spatial information, and coherence issues, this groundwork is crucial for future research.
\vspace{-5pt}
\paragraph{Online Visual Dialogue.} Existing research in visual understanding~\cite{Large-scale, Vd-bert, Video-llama, Videochat, Moviechat, Video-chatgpt, Deepspeed-visualchat} primarily focuses on enabling the input of complete images or videos into Multimodal Large Language Models (MLLMs)~\cite{blip2, llava, minigpt4, mplug-owl2, mplug-owl3, blip-instruction} for offline text-based multi-turn dialogues. While some works~\cite{Visual-chatgpt, Llava-next-interleave, Interleaved1, Interleaved2, Interleaved3} have extended this dialogue paradigm to interleaved text and images, and some online video understanding methods~\cite{Streaming, Flash-VStream, Streaming-long} have advanced the processing of continuous video streams, they still fall short of enabling online video dialogue. In particular, these methods fail to provide real-time, proactive responses within video streams. Videollm-online~\cite{VideoLLM-online} introduces a new paradigm for online video dialogue, establishing the foundational LIVE framework for video assistants. However, LIVE still faces challenges in accurate response determination, precision response generation, as well as a balanced trade-off between efficacy and efficiency.

% \textbf{The Fast \& Slow Concept.} The "Fast \& Slow" framework, derived from Daniel Kahneman's "Thinking, Fast and Slow,"~\cite{thinking} delineates two cognitive systems: the rapid, intuitive "System 1" and the deliberate, rational "System 2"~\cite{Towards-understanding, Visual-cot, Visual-agents, fei2024video}. SlowFast-LLaVA~\cite{Slowfast-llava} architecture integrates low-frame-rate fine-grained pooled tokens with high-frame-rate coarse-grained tokens, aggregating features sampled from video frames. In complex queries, FaST~\cite{Visual-agents} selectively activates a slow-thinking pathway for complex visual queries, enhancing image detail by identifying salient areas in visual inputs. Inspired by this, our video assistant swiftly navigates video streams to identify response moments while ensuring careful consideration for accurate and coherent answers.
\vspace{-10pt}
\paragraph{The Fast and Slow Concept.} Originating from Daniel Kahneman's ``\textit{Thinking, Fast and Slow}"~\cite{thinking}, which distinguishes between two cognitive systems: the fast, intuitive``System 1" and the deliberate, rational``System 2"~\cite{Towards-understanding, Visual-cot, Visual-agents, fei2024video}. Many works have utilized the ``Fast and Slow" mechanism, such as SlowFast-LLaVA~\cite{Slowfast-llava}, which performs pooling at different granularities on videos with varying frame rates to enrich features. FaST~\cite{Visual-agents} selectively activates ``System 2" for complex visual queries to enhance detailed image understanding. SLOWFAST-VGEN~\cite{SlowFast-VGen} leverages a slow-fast learning loop for action-driven long video generation. Inspired by these works and dual-path approaches~\cite{shao2022open, cat, lion}, we apply the ``Fast and Slow" mechanism to our online video assistant, decoupling simple response determination from complex response generation to enhance efficacy and efficiency.

\vspace{-10pt}
\paragraph{Routing-Based Modeling.} The Mixture of Experts mechanism~\cite{moe, lion, mome, moe2} is widely adopted in multimodal learning~\cite{shao2023detecting, shao2024detecting, shao2019multi, Optimus-1, Optimus-2, chen2024spa} within current large multimodal language models. It explores routing schemes between expert networks across different knowledge domains, selecting specific experts based on task types. Mixture-of-Depths mechanism~\cite{mod, Gamma-MOD, videollm-mod} investigates routing schemes for tokens involved in computation within transformer layers, reducing excessive computational overhead. We apply and enhance these two mechanisms in online video dialogue, enabling dynamic aggregation of diverse visual tokens while adaptively discarding redundant visual tokens.

%% file: sec/3_method.tex
\vspace{-5pt}
\section{LION-FS}

% 1. slow path 部份需要呈现确定的结果，等实验出来后调整
% [SOLVED] 2. 只讲了怎么训练，缺少一个pipeline介绍：先用fastpath判断，再用slowpath生成
% 3. fastpath 能否使用和slowpath相同的画法（参考SlowFast-LlaVA）

% In this section, we introduce the LION-FS framework, which optimizes the decision-making and generation of responses through two paths: Fast Path and Slow Path. This decoupling strategy aligns with human cognitive processes, enhancing the performance and efficiency of video assistant while also improving its perceptual capabilities.
\vspace{-5pt}
The whole framework of LION-FS is shown in Figure~\ref{fig:framework}. It divides the online video dialogue into Fast Path and Slow Path for response determination and generation, respectively. This decoupling strategy aligns with human dialogue processes and excels in both efficacy and efficiency. 

\vspace{-5pt}
\subsection{Online Video Dialogue Modeling}
% The online video dialogue task requires the assistant to engage in conversations with the user based on a continuously updated video stream. The dialogue pipeline could be divided into two main phases: response determination and generation. For each incoming video frame, the assistant first determines whether it is appropriate to answer the question at that moment. If a response is needed, the assistant then generates an answer auto-regressively based on the previous video context.
\vspace{-5pt}
The online video dialogue task requires the assistant to engage with users through continuously updating video streams. For each incoming frame, the assistant first determines whether an immediate response is appropriate. If a response is warranted, it generates an answer auto-regressively based on the preceding video context.

% The target of the two phrases can be expressed by
% % \begin{equation}
% \begin{align}
% & \max P([EOS] \mid [Ctx]), \\
% & \max P([Txt]_{j+1} \mid [Ctx], [Frm]_{k}, [Txt]_{\leq j}) \label{eq:5}
% \end{align}
% % \end{equation}
% where $[Ctx]$ represents the previous vision-language content within the current LLM window, including historical user queries, video frames, and assistant responses. The response decision-making process can be interpreted as the prediction of the EOS token.

We adopt a training approach aligned with LIVE~\cite{VideoLLM-online}, where the objective consists of two components: a streaming loss and a standard language modeling (LM) loss, corresponding to response determination and response generation, respectively. The combined loss is formulated as:
\begin{equation}
\hspace{-0.7em} 
\text{Loss} = \frac{1}{N} \sum_{j=1}^{N} ( \underbrace{- w s_j \log P_j^{\texttt{[EOS]}}}_{\text{Streaming~Loss}} \underbrace{-l_{j+1} \log P_j^{\texttt{[Txt]}_{j+1}}}_{\text{LM~Loss}} )
% \hspace{-1em} 
\end{equation}
Here, $ P_j^\texttt{[EOS]} $ represents the probability that the LLM predicts the End-Of-Sequence (EOS) token\footnote{The EOS token can be replaced by any other token, serving the purpose of signaling whether a response should be generated.} at the $j$-th token. $ P_j^{\texttt{[Txt]}_{j+1}} $ denotes probability of autoregressively predicting the ($j$+$1$)-th text token within the response. The weight-balancing parameter $w$ has a default value of $1$. 
The binary condition coefficients $ s_j $ and $l_{j+1} $, consistent with the LIVE setting, take the value of 1 during response determination and generation, respectively.

\subsection{Fast Path: Routing-Based Response Determination}

% The Fast Path of LION-FS is responsible for response determination, which requires the assistant to process video frames at a high rate but does not require much language-generating capability. To achieve this, we design an image-video dual encoding framework to increase frame rate and a sparsely activated LLM to reduce latency, as depicted in the Fast Path of Figure~\ref{fig:framework}.
To capture temporal information from high-frame-rate video streams and mitigate the mismatch between first-person tasks and third-person pre-trained visual features, we propose Token Aggregation Router and  Token Dropping Router as the Fast Path. This framework adaptively aggregates distinct features while discarding redundant ones, as illustrated in the Fast Path of Figure~\ref{fig:framework}.

\vspace{-10pt}
\paragraph{Dual Encoding with Token Aggregation Router.}
Current online VideoLLMs~\cite{VideoLLM-online, videollm-mod} utilize a general image encoder to process video frames separately and rely solely on the LLM to interpret videos with frame-rate downsampled features, which presents two significant limitations.
Firstly, the frame-by-frame processing approach significantly limits the frame rate due to the computational complexity of LLM. This limitation makes it difficult to capture inter-frame variations in perspective and action, particularly in the downsampled feature maps. 
Secondly, since general image encoders are trained on third-person images, they cannot understand first-person scenes effectively due to the perspective discrepancy.
% Firstly, the frame-by-frame processing approach in online tasks greatly restricts the frame rate, resulting in difficulty capturing inter-frame variations in perspective and action on the downsampled feature maps. Secondly, there exists a substantial discrepancy between the pre-training data of general image encoders and first-person video frames, leading to a mismatch between training and inference data. However,~\cite{Freeva, Long-context, Llava-onevision} have demonstrated that general image encoders possess rich visual knowledge and generalization capabilities, enabling them to perform effectively in video tasks. Therefore, we propose a Routing-Based Image-Video Dual Encoding approach. 
% Secondly, there exists a substantial discrepancy between the training data of general image encoders and first-person video frames, leading to insufficient first-person information.

% To address these limitations, we propose to introduce an additional video encoder pre-trained on egocentric data to process video frames in groups at a higher rate. Such a group-by-group encoding mechanism enables vision encoders to capture fine-grained temporal information while enhancing efficiency.Moreover, it has been demonstrated~\cite{Freeva, Long-context, Llava-onevision} that general image encoders possess rich visual knowledge and generalization capabilities, which is not dispensable. Therefore, we propose a Routing-Based Image-Video Dual Encoding approach to utilize information from the two encoders adaptively and efficiently.
To address these limitations, we introduce an additional video encoder pre-trained on egocentric data to process video frames in groups at a higher rate. Such a group-by-group encoding approach captures fine-grained temporal information while ensuring real-time video processing and enriching visual features with first-person knowledge. Moreover, prior studies~\cite{Freeva, Long-context, Llava-onevision} have shown that general image encoders possess rich visual knowledge and strong generalization capabilities, which are indispensable. Therefore, we propose a Dual Encoding with Token Aggregation Router that temporally aligns the two types of visual features, adaptively and efficiently leveraging both encoders.

Specifically, we employ EgoVLPv2~\cite{egovlpv2} as the video encoder $E_{ego}$ to process 8 FPS video streams by grouping every 4 frames. Meanwhile, we use SigLIP~\cite{siglip} as the image encoder $E_{gen}$ to process 2 FPS video streams. Each 0.5-second segment of video is encoded into two sequences containing either 1 or 10 tokens (including 1 CLS token and 3×3 pooled tokens). The process is formulated as follows:

{
\fontsize{9}{1}\selectfont
% \small
\begin{align}
% \hspace{-2em}
\texttt{[Frm}_s\texttt{]}_{i} &= E_{gen}(\text{Frm}_{i}), \\
\texttt{[Frm}_t\texttt{]}_{i} &= E_{ego}(\text{Frm}_{i}, \text{Frm}_{i}^0, \text{Frm}_{i}^1, \text{Frm}_{i}^2)
% \hspace{-1em}
\end{align}
}
% \begin{align}
% % \hspace{-2em}
% \text{[Frm}_s\text{]}_{i} &= E_{gen}(\text{Frm}_{i}), \\
% \text{[Frm}_t\text{]}_{i} &= E_{ego}(\text{Frm}_{i}, \text{Frm}_{i}^0, \text{Frm}_{i}^1, \text{Frm}_{i}^2)
% % \hspace{-1em}
% \end{align}
\hspace{-0.3em}where $\texttt{[Frm}_s\texttt{]}_{i}$ and $\texttt{[Frm}_t\texttt{]}_{i}$ represent the general image tokens from $E_{gen}$ and the first-person group tokens from $E_{ego}$, respectively. $\text{Frm}_{i}$ denotes the frame sampled at 2 FPS, while $ \text{Frm}_{i}^0 $, $ \text{Frm}_{i}^1 $, and $ \text{Frm}_{i}^2 $ are frames uniformly sampled from 8 FPS frames surrounding $\text{Frm}_{i}$.

\begin{figure}[t]
  \centering
   \includegraphics[width=1\linewidth]{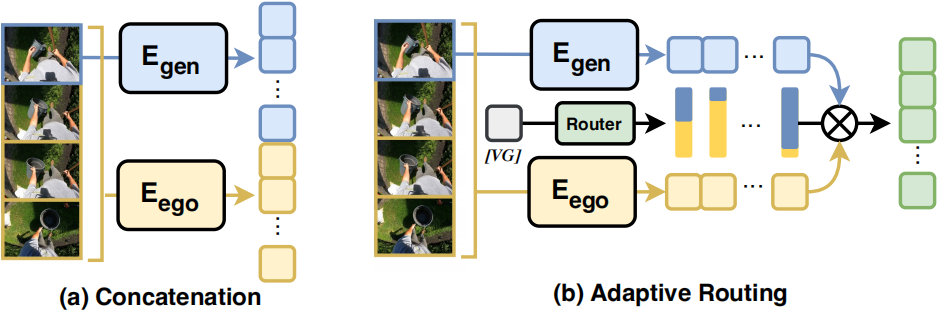}
   \caption{Different Token Aggregation Strategies: (a) Concatenate tokens along the sequence dimension. (b) Aggregate tokens based on adaptive weights generated by the router to perform customized routing. It can achieve visual information aggregation without increasing token numbers.}
    \vspace{-10pt}
   \label{fig:routing}
\end{figure}

As shown in Figure~\ref{fig:routing}(a), simply concatenating $ \texttt{[Frm}_s\texttt{]}_{i} $ and $ \texttt{[Frm}_t\texttt{]}_{i} $ provides a straightforward approach to enriching visual information. However, it increases the token sequence length, negatively impacting the LLM's decoding efficiency. 
% We also observe and validate the unique advantages of different token types (Section~\ref{sec431}).
To further balance efficacy and efficiency, as shown in Figure~\ref{fig:routing}(b), we propose the adaptive routing method to aggregates $ \texttt{[Frm}_s\texttt{]}_{i} $ and $ \texttt{[Frm}_t\texttt{]}_{i} $ adaptively based on weights generated by the Token Aggregation Router. 
% Specifically, the Token Aggregation Router generates a customized weight for each token, followed by a weighted sum to aggregate them. Due to the high variability in the video frame content, there is no one-fits-all fusion strategy.
Specifically, we consider that a large-scale pre-training allows SigLIP to effectively and comprehensively capture both visual information and frame-to-frame variations. It can be utilized as strong guidance for customized routing. Based on this consideration, the CLS token from $ \texttt{[Frm}_s\texttt{]}_{i} $ is used as Visual Guidance $\texttt{[VG]}$ as the condition of the router.
% from $ \texttt{[Frm}_s\texttt{]}_{i} $ and $ \texttt{[Frm}_t\texttt{]}_{i} $, maximizing the advantages of each token. 
% To generate a 
% Meanwhile, we consider that CLS token extracted by SigLIP possesses rich general knowledge and strong generalization capability, allowing it to comprehensively capture both visual information and frame-to-frame variations. Based on this consideration, the CLS token from $ \texttt{[Frm}_s\texttt{]}_{i} $ is used as Visual Guidance $\texttt{[VG]}$ for feature allocation. 
The whole process can be formulated as follows:

{
\fontsize{8}{1}\selectfont
% \small
\begin{align}
    & G_f(\texttt{[VG]}) = \text{SoftMax}(W_2 (\sigma(W_1 \texttt{[VG]} + b_1)) + b_2), \\
    & {\texttt{[Frm]}_i} = G_f(\texttt{[VG]})_0 \times \texttt{[Frm}_s\texttt{]}_{i} + G_f(\texttt{[VG]})_1 \times \texttt{[Frm}_t\texttt{]}_{i}
\end{align} 
}
% \hspace{-0.3em}Through the router $G_f$, we generate adaptive aggregation weights for both types of visual features based on $\texttt{[VG]}$, resulting in ${\texttt{[Frm]}_i}$. ${\texttt{[Frm]}_i}$ integrates third-person general spatial features with first-person dense temporal features, significantly enhancing the visual information without increasing the number of tokens.
\hspace{-0.4em}Through the router $G_f$, we generate adaptive aggregation weights for both types of visual features based on $\texttt{[VG]}$, resulting in ${\texttt{[Frm]}_i}$.
% The selected $\texttt{[VG]}$ comes from the CLS token extracted by SigLIP, which possesses rich general knowledge and strong generalization capability, allowing it to comprehensively capture both visual information and frame-to-frame variations. As a result, the CLS token extracted by SigLIP outperforms both the CLS token from EgoVLPv2 and the query text token that is consistent across all frames. 
Ultimately, ${\texttt{[Frm]}_i}$ achieves the adaptive fusion of third-person general spatial features with first-person dense temporal features, significantly enhancing visual information without increasing output token numbers.

% We have enhanced the information content of visual tokens while also needing to discard redundant ones. Taking Figure~\ref{fig:routing}(c) as an example, $[Frm_s]_{(i, 0)}$ is the CLS token, capturing rich features that should be retained, while $[Frm_s]_{(i, 1:10)}$ represents spatially pooled features of the entire frame.

\vspace{-10pt}
\paragraph{Sparse Decoding with Token Dropping Router.}
\label{Token-Dropping-Router}
In addition to aggregating temporal and egocentric vision features, the online assistant has to be efficient enough to accept real-time video frames.

% 1. 视觉信息冗余度高无需全部保留 2. 决策无需dense llm
% 我们提出MoD进行....

% Although the Routing-Based Image-Video Dual Encoding framework adds some fine-grained visual information, in first-person scenes, our focus is on interaction areas between people and the environment, making the redundancy of spatial features more pronounced when frame variations are minimal. We employ the following routing to discard redundant spatial pooling tokens:

Although the Routing-Based Image-Video Dual Encoding framework enriches diverse visual information, visual feature redundancy remains inevitable. This redundancy arises from two main reasons: (i) we use average pooling tokens overall patch tokens for each frame, while in first-person scenes, the focus is typically on interaction areas between individuals and the environment; (ii) minimal variation between some consecutive frames leads to repeated, highly similar visual tokens. Therefore, we adopt the following routing strategy to drop redundant tokens:

{
\fontsize{9}{1}\selectfont
% \small
\begin{equation}
    \begin{aligned}
        \hspace{-1em}
        \texttt{[Frm]}_{(i, n)}^{l+1} = \begin{cases} 
        r_{(i, n)}^l f_i(\tilde{X}^l) + \texttt{[Frm]}_{(i, n)}^{l}, & \text{if } r_{(i, n)}^l > P_\beta^l \\
        \texttt{[Frm]}_{(i, n)}^{l}, & \text{if } r_{(i, n)}^l < P_\beta^l
        \end{cases}
    \end{aligned}
    \hspace{-1em} 
\end{equation}
}
\hspace{-0.4em}where $0 \mathbin{\le} n \mathbin{\le} 9$ indicates that redundant tokens can be selectively discarded from the 10 tokens of each frame. $ r_{(i, n)}^l = w_\theta^T \texttt{[Frm]}_{(i, n)}^{l} $ represents the routing weight scalar generated via linear projection for each token at the $ l $-th transformer layer. Only tokens with weights exceeding $P_\beta^l$ are retained at this layer. $\beta$ denotes the user-defined proportion of discarded tokens relative to the total number of visual tokens, and $ P_{\beta}^l $ is the $\beta$-th percentile of the routing weight set $ r^l $ at layer $ l $. $ f_i $ represents the self-attention and FFN operations at the current layer, meaning that only the top-k weighted visual tokens, filtered by their routing weights, participate in the interleaved vision-text dialogue through $ f_i $, defined as $ \tilde{X}^l = {\text{Interleaved}(\texttt{[Frm]}_{(i, \text{top-k}[0:10])}^{l}, \texttt{[Txt]}_{j}^{l}) \mid \forall i, j } )$.

Through the proposed Routing-Based Fast Path, we achieve a fourfold increase in frame rate compared to LIVE for real-time video streaming, while enriching dense temporal information and leveraging pre-trained first-person features, resulting in more accurate and efficient response determination.

\begin{table*}[t]
\caption{\textbf{Main Results.} ``\dag'' indicates that the comparison model was not trained on the evaluation dataset, and we retrained it for a fair comparison. Since the code for VideoLLM-MOD has not been released, the experimental results were reproduced based on its paper. ``*'' indicates that the experimental data comes from the results reported in the original paper, where our model was trained and evaluated using the same experimental settings. The experimental results show that LION-FS outperforms existing comparison methods on most metrics, demonstrating its strong ability on video-stream online dialogue tasks.}
    \label{tab:main_results}
    \vspace{-5pt}
    \centering
    \footnotesize
    % \scriptsize
    % \setlength{\tabcolsep}{3pt}
    \begin{tabular}{l|cccc}
    \toprule
    % \textbf{Method} & \multicolumn{4}{c|}{\textbf{Ego4D/Ego-Exo4D Narration Validation}} & & \\
    \textbf{Method} & \textbf{LL-PPL} $\downarrow$ & \textbf{TimeDiff} $\downarrow$ & \textbf{Fluency} $\uparrow$ & \textbf{LM-Correctness} $\uparrow$  \\
    % \midrule
    \hline
    \rowcolor[HTML]{EFEFEF} \multicolumn{5}{c}{\textit{Ego-Exo4D Narration Validation}} \\
    \hline
    VideoLLM-online\dag ~\cite{VideoLLM-online} & 2.24 &0.78 &33.7\% &44.8\%   \\
    VideoLLM-MoD\dag~\cite{videollm-mod} &2.12 &0.82 &33.8\% &45.3\%   \\
    \rowcolor[HTML]{faf0e6}
    LION-FS &\textbf{2.04} &\textbf{0.74} &\textbf{36.5\%} &\textbf{48.2\%}  \\
    \hline
    \rowcolor[HTML]{EFEFEF} \multicolumn{5}{c}{\textit{Ego4D Narration Validation}} \\
    \hline
    VideoLLM-online* ~\cite{VideoLLM-online} &2.40 &\textbf{2.04} &45.3\% & 49.0\%   \\
    VideoLLM-MoD*~\cite{videollm-mod}  &2.41 &\textbf{2.04} &45.2\% &48.9\%   \\
    \rowcolor[HTML]{faf0e6}
    LION-FS &\textbf{2.09} &2.15 &\textbf{46.1\%} &\textbf{52.4\%}  \\
    \bottomrule
    \end{tabular}
\end{table*}

\subsection{Slow Path: Multi-granularity Keyframe Augmentation}

% In the Fast Path, we treat each frame equally, enabling accurate and efficient determination of event occurrences.
% However, the generation phase prioritizes high-quality visual information. Representing a frame with 10 tokens limits the ability to capture finer features and fails to highlight critical interaction areas between individuals and the environment in first-person scenarios. As indicated in Equ.\eqref{eq:5}, the model determines the need for response occurrence at keyframes in an autoregressive manner. Thus, these decisive frames, which signify the transition of events or actions in the video, should not be treated the same as frames deemed silent. Consequently, we perform multi-granularity expansion on keyframes, establishing the training-free Slow Path.
In the Fast Path, we treat each frame equally to enable accurate and efficient response determination. However, representing a frame with only $10$ tokens tends to lose fine-grained visual information, which degenerates video-language fusion and thus affects precise response generation. Additionally, adding fine-grained features to all frames is impractical, as it would significantly impact the assistant's efficiency, making real-time video streaming infeasible. Therefore, we propose a training-free Slow Path for multi-granularity keyframe augmentation. We define the current frame, at which response is determined, as the keyframe, serving as a transition point for events or actions in the video. We apply both global uniform augmentation and local adaptive augmentation to these keyframes, resulting in Grid tokens and Box tokens.

\vspace{-10pt}
% \paragraph{Grid Tokens \& Box Tokens in Thinking Template.}
\paragraph{Global uniform augmentation for Grid Tokens.}
% In supplementing fine-grained tokens, we leverage the LLM's strong contextual modeling capabilities while adhering to the 10-tokens pattern established during training for each frame. Specifically, we divide the keyframe into four grids, applying the same $3 \times 3$ pooling used for ordinary frames, and introduce inter-frame separators between each grid. This approach treats the keyframe as four distinct frames to enhance fine-grained information. The continuous visual feature representation of a keyframe is given by: ``[$\textless CLS \textgreater , \textless Frame:coarse-grained\textgreater , \textless Grid:fine-grained\textgreater \times 4$]''. 
To supplement global fine-grained tokens and fully leverage the 10-token representation per frame established by Fast Path, we apply a unified grid-based strategy to all keyframes. Specifically, each keyframe is divided into four uniform grids, with each grid subjected to the same 3×3 pooling operation used for non-keyframes. In addition, inter-frame separators are introduced to delineate pooled features between grids, resulting in a transformation from 1×6×6 to 4×3×3 pooled tokens. This process enables each keyframe to be treated as four distinct sub-frames that are sequentially input to the LLM, thereby enhancing the granularity of information captured in a train-free manner, while ensuring that fine-grained spatial details from each region are preserved and effectively represented.

\vspace{-10pt}
\paragraph{Local adaptive augmentation for Box Tokens.}
% In first-person scenarios, video assistant primarily focuses on the interaction zones between humans and environment, which are critical for actions and events. We introduce box tokens to guide the assistant's attention through target localization in these core areas. Specifically, we employ Faster R-CNN~\cite{Faster-R-CNN} to detect hand positions by minimizing the distance and squared error between hand and object bounding boxes~\cite{hands-objects, Detecting}, followed by NMS optimization to regress the final bounding boxes of objects interacting with the hands. Using these bounding boxes, we spatially filter the unpooled 576 patch tokens within each bounding box and perform global pooling separately to obtain a single token representation , namely the hands tokens and objects tokens.To ensure the LLM attends to these box tokens, we incorporate prompt guidance, such as ``$[Please~ focus~ on~ \textless hands \textgreater ~and~ \textless objects \textgreater]$''. 
In first-person scenarios, video assistants predominantly focus on interaction regions between humans and the environment, which are critical for capturing actions and events. To achieve this, we introduce box tokens to direct the assistant's attention toward these key regions through object localization. Specifically, (i) Faster R-CNN~\cite{Faster-R-CNN} is employed to detect hand positions, followed by refining the bounding boxes of objects interacting with the hands by minimizing the distance and squared error between hand and object anchor boxes~\cite{hands-objects, Detecting}, along with NMS optimization. (ii) Based on these bounding box coordinates, we identify the corresponding tokens from the unpooled 576 patch tokens and perform global pooling within each bounding box to derive single-token representations for hands and objects, collectively forming the box tokens.

% We integrate the results of multi-granularity keyframe augmentation to establish a Thinking Template within interleaved frame-text dialogues, as illustrated in Figure~\ref{fig:framework}. The specific format is as follows: ``[$\textless CLS \textgreater , \textless Frame \textgreater , \textless Grid \textgreater \times 4][ User: Please~ focus~ on~ \textless hands \textgreater ~and~ \textless objects \textgreater. ~Assistant:$]''. 
After performing multi-granularity keyframe augmentation, we obtained the Grid Tokens and Box Tokens for each keyframe. We integrated these tokens into the interleaved frame-text dialogue, constructing a Multimodal Thinking Template, as depicted in Figure~\ref{fig:framework}. The specific format is as follows: \textbf{``\textit{Stream: \texttt{[Frame Tokens] [Grid Tokens]} User: Please focus on \texttt{[Box Tokens]}. Assistant: "}}. The process is summarized as follows: when the Slow Path receives a response indication (predicting ``\textit{Assistant:}'') from the Fast Path, the current frame is deemed a keyframe. Multi-granularity augmentation is then applied to this keyframe to generate the Multimodal Thinking Template, which serves as a customized multimodal prompt, replacing ``\textit{Assistant:}''. 
% This training-free approach enhances the flexibility of the assistant's responses, avoiding bias imposed by brief atomic action descriptions in the training data, and improving the precision of them.
This seamless insertion of a Multimodal Thinking Template into online video dialogues enriches multi-granularity visual information in a training-free manner. It allows responses to move beyond the limited brief atomic action descriptions in the training data, thus improving the precision of the responses.

%% file: sec/4_experience.tex
\section{Experience}

\begin{table*}[t]
\caption{\textbf{Ablation Study on Token Aggregation Router.} $E_{\text{gen}}$ and $E_{\text{ego}}$ represent the number of tokens from SigLIP and EgoVLPv2, respectively, while ``Fusion" denotes the token count obtained through the fusion strategy. The results indicate adaptive routing aggregation improves visual integration and captures temporal-spatial correlations, significantly boosting model efficacy.}
    \label{tab:ablation_moe}
    \vspace{-5pt}
    \centering
    \footnotesize
    \begin{tabular}{c|ccc|cccc}
    \toprule
    \textbf{Method} & \multicolumn{3}{c|}{\textbf{Token Number}} & \multicolumn{4}{c}{\textbf{Ego-Exo4D Narration Validation}} \\ Aggregation strategy & $E_{\text{gen}}$ & $E_{\text{ego}}$ & Fusion & \textbf{LL-PPL} $\downarrow$ & \textbf{TimeDiff} $\downarrow$ & \textbf{Fluency} $\uparrow$ & \textbf{LM-Correctness} $\uparrow$ \\
    \midrule
    - & 10         & - & 10 & 2.24 & 0.78 & 33.7\% & 44.8\%  \\
    - & - & 10 & 10 &  2.29   &    1.05    &    36.8\%    &   47.8\%  \\
    \midrule
     & 10 & 10 & 20 & 2.25 & 1.65 & 27.7\% & 45.8\% \\
    Concatenation & 10 & 1 & 11 & 2.29 & 0.71 & 35.8\% & 45.2\% \\
     & 1 & 10 & 11 & 2.42  &  1.07 & 37.1\% &47.6\%   \\
    \midrule
     & 10 & 10 & 10 & 2.25 & 0.75 & 34.7\% & 44.7\%   \\
    Addition & 10 & 1 & 10 & \textbf{2.18} & 0.71 & 36.2\% & 46.9\%   \\
     & 1 & 10 & 10 &   2.38   &   1.05&   33.8\%   & 45.0\%        \\
    \midrule
    Learnable Weighting & 10 & 10 & 10 &2.35   &0.74  &34.7\%  &45.6\%   \\
    \rowcolor[HTML]{faf0e6}
    Adaptive Routing & 10 & 10 & 10 &2.25   &\textbf{0.67}  &\textbf{38.1\%}  &\textbf{48.0\%}  \\
    \bottomrule
    \end{tabular}
    \vspace{-10pt}
\end{table*}

\subsection{Experimental Settings}
All experiments can be completed using 80G A800 GPU. Please refer to Appendix for
implementation details and training details.

\vspace{-10pt}
\paragraph{Datasets.}
In an online setting, we validated the effectiveness of our proposed LION-FS model using the egocentric video datasets, Ego4D and Ego-Exo4D.
% \begin{itemize}[label=$\bullet$]
%     \item \textbf{Ego4D Narration Stream Benchmark}~\cite{ego4d}: According to  VideoLLM-online~\cite{VideoLLM-online} , we use dense Ego4D timestamped-narrations to create a streaming dataset, aiming to generate narrations in a timely manner, akin to those produced by human annotators in Ego4D.
%     \item \textbf{EgoExo4D Benchmark}~\cite{Ego-exo4d}: Ego-Exo4D is a multiview, temporally aligned video dataset (including first-person view). We performed the same operations on this dataset as we did with Ego4D. Due to its smaller scale compared to Ego4D, we used it for ablation experiments to quickly evaluate the effectiveness of different model modules.
% \end{itemize}
\textbf{Ego4D Narration Stream Benchmark}~\cite{ego4d}: According to  VideoLLM-online~\cite{VideoLLM-online}, we use dense Ego4D timestamped-narrations to create a streaming dataset, aiming to generate narrations in a timely manner, akin to those produced by human annotators in Ego4D.
\textbf{Ego-Exo4D Benchmark}~\cite{Ego-exo4d}: Ego-Exo4D is a multiview, temporally aligned video dataset including first-person view. We perform the same operations on this dataset as with Ego4D. Due to its smaller scale compared to Ego4D, we use it for ablation experiments to quickly evaluate the effectiveness of different model modules.

\vspace{-10pt}
\paragraph{Evaluation metrics.}
For the online benchmarking, following the setup of VideoLLM-online~\cite{VideoLLM-online}, we use Language Modeling Perplexity (LM-PPL) and LM-Correctness to evaluate the language modeling capability of our LION-FS model at specific timestamps. To assess the model's temporal alignment ability, we use Time Difference (TimeDiff) and Fluency to comprehensively measure the quality of language modeling and temporal effectiveness.

\subsection{Main Results}

\begin{table*}[t]
\caption{\textbf{Ablation Study on Token Dropping Router.} ``No Dropping" represents the best result of Adaptive Routing Aggregation without using the Token Dropping Router. ``Random Dropping" refers to adding a non-trainable Token Dropping Router on top of Adaptive Routing Aggregation. $\beta$ indicates the dropout rate of visual tokens. The Dropping Layers experiment was conducted under the condition of  ${\beta} $ = 0.5, while the Dropping Ratio experiment was performed under the ``Interleaved Layers" setting. The experimental results show that the ``Interleaved Layers"  $\&~ {\beta} $ = 0.5 routing configuration strikes a good balance between model performance and efficiency.}
    \label{tab:ablition_mod}
    \vspace{-5pt}
    \centering
    \footnotesize
    \begin{tabular}{c|cccc|cc}
    \toprule
    \textbf{Method} & \multicolumn{4}{c|}{\textbf{Ego-Exo4D Narration Validation}} & \textbf{FLOPs} & \textbf{Training Cost \& Speedup}\\ Dropping strategy & \textbf{LL-PPL} $\downarrow$ & \textbf{TimeDiff} $\downarrow$ & \textbf{Fluency} $\uparrow$ & \textbf{LM-Correctness} $\uparrow$ & &  \\
    \hline
    No Dropping & 2.25 & 0.67 & 38.1\% & 48.0\% &61.44T  &  6.4h\& n/a\\
    Random Dropping &15.48 &1.93 &20.7\% &30.3\% &- &- \\
    % \rowcolor[HTML]{faf0e6}
    % Route Dropping & & & & & & \\
    \rowcolor[HTML]{faf0e6}
    \hline
    \rowcolor[HTML]{EFEFEF}
    \multicolumn{7}{c}{\textbf{Dropping Layers}} \\
    \hline
    All Layers &2.18  & 0.80  &34.1\%  &45.5\%  &41.39T  &5.0h \& 1.28 $\times$  \\
    Deep Layers &2.15  &0.77  &35.6\%  &46.4\%  &48.89T  &5.4h  \& 1.18 $\times$\\
    \rowcolor[HTML]{faf0e6}
    Interleaved Layers &2.16  & 0.74  & 36.5\%  &47.0\%  &51.40T  &5.7h \& 1.12 $\times$  \\
    Interleaved \& Deep Layers &2.13  &0.80  &33.9\%  &45.1\%  &45.14T  &5.3h \& 1.21 $\times$ \\
    \hline
    \rowcolor[HTML]{EFEFEF} \multicolumn{7}{c}{\textbf{Dropping Ratio}} \\
    \hline
    $\beta = 0.2$ &2.21 &0.73 &36.5\% &46.7\% &57.44T &6.0h \& 1.05$\times$ \\
    \rowcolor[HTML]{faf0e6}
    $\beta = 0.5$ &2.16 &0.74 &36.5\% &47.0\% &51.40T &5.7h  \& 1.12 $\times$ \\
    $\beta = 0.8$ & 2.28 & 1.10 & 35.9\% & 46.8\% & 45.37T & 5.4h \& 1.18 $\times$ \\
    % $\beta = 0.9$ & & & & & & \\
    \bottomrule
    \end{tabular}
    \vspace{-10pt}
\end{table*}

\begin{table}[t]
\caption{\textbf{Ablation Study on Adaptive Routing for Visual Guidance $\texttt{[VG]}$ Selection.} The results demonstrate that the CLS token proposed by SigLIP, with its rich knowledge and generalization ability, most effectively optimizes the weights between visual tokens, achieving superior efficacy.}
    \label{tab:main_results_vg}
    \vspace{-5pt}
    \centering
    \scriptsize
    \setlength{\tabcolsep}{3pt}
    \begin{tabular}{c|cccc}
    \toprule
    \textbf{$\texttt{[VG]}$ Source} & \textbf{LL-PPL} $\downarrow$ & \textbf{TimeDiff} $\downarrow$ & \textbf{Fluency} $\uparrow$ & \textbf{LM-Correctness} $\uparrow$  \\
    \midrule
    \rowcolor[HTML]{faf0e6}
    SigLIP  & 2.25 &0.64 &38.1\% &48.0\%  \\
    EgoVLPv2  &2.45 &0.70 &38.1\% &47.7\%   \\
    SigLIP \& EgoVLPv2  &2.40 &1.08 &37.8\% &47.7\%   \\
    \bottomrule
    \end{tabular}
    \vspace{-10pt}
\end{table}

We compare our method with existing video streaming dialogue methods~\cite{VideoLLM-online, videollm-mod} on the Ego4D and Ego-Exo4D narration stream benchmark, the results of which are summarised in Table~\ref{tab:main_results}.
LION-FS consistently outperforms other methods, particularly in Fluency and LM-Correctness metrics, showcasing its advanced capabilities in language modeling and temporal alignment. 
While it slightly underperforms on TimeDiff in Ego4D Narration Validation, due to the short average response length of 6.73 words. In contrast, it outperforms all metrics in the Ego-Exo4D Narration Validation (10.96 words), highlighting its focus on balancing response determination and generation.
Additionally, LION-FS can process video streams at a higher frame rate (four times that of ~\cite{VideoLLM-online, videollm-mod}), achieving greater efficiency.
The dual advantages in efficacy and efficiency are brought by the novel fast \& slow two-path optimization scheme.

\subsection{Ablation Study}

% \begin{figure*}[t]
%   \centering
%    \includegraphics[width=1.0\linewidth]{fig/mod.png}
%    \caption{mod}
%    \label{fig:mod}
% \end{figure*}

\subsubsection{Analysis on Token Aggregation Router}
\label{sec431}

Table~\ref{tab:ablation_moe} presents the ablation study results on visual token aggregation strategies. Experiments show that SigLIP outperforms EgoVLPv2 in LL-PPL and TimeDiff, while EgoVLPv2 excels in Fluency and LM-Correctness. These findings highlight the limitations of using on a single visual encoder, as it cannot provide comprehensive visual information. Features from two different encoders can complement each other, offering a more complete representation.
% We first performed experiments using SigLIP and EgoVLPv2 separately as visual encoder. The results show that SigLIP outperforms EgoVLPv2 in LL-PPL and TimeDiff metrics, while EgoVLPv2 excels in Fluency and LM-Correctness. Single visual encoder can not provide comprehensive information and  has its inherent limitations. 
% The evidence suggests that using a single visual encoder has its inherent limitations. We assume that SigLIP, as a general visual encoder, is able to capture rich spatial visual information, which helps reduce the perplexity of the LLM and makes the generated content closer to the training distribution. In contrast, EgoVLPv2, as a first-person visual encoder, not only extracts unique first-person visual information but also effectively utilizes temporal information, thereby enhancing the coherence and correctness of the generated text.

% We believe that using a single-view visual encoder has its limitations, so we subsequently introduced different aggregation strategies to integrate information from both first-person and third-person views, thereby improving overall performance.

To aggregate features from two encoders, we first simply concatenate different features to provide the model with rich visual information. However, this approach cannot achieve consistent improvement while impacting the LLM's decoding efficiency due to the increase in the token length. Then, we attempt to directly add the visual tokens from different encoders to prevent the token length increase. 
We separately perform the ``10+10" (where each token is added individually) and ``10+1/1+10" (where the CLS token from one encoder is added to the CLS token from the other encoder) aggregation strategies. It is evident that the experimental results of ``10+1/1+10" outperform those of ``10 + 10" in most metrics, leading us to believe that different visual tokens require different addition strategies. 
Based on these findings, we propose the Token Aggregation Router, which assigns adaptive weights to each token guided by Visual Guidance $\texttt{[VG]}$. Table~\ref{tab:main_results_vg} shows that using the CLS token from SigLIP as $\texttt{[VG]}$ outperforms the CLS token from EgoVLPv2 or their combination. The experimental results demonstrate that the adaptive routing method surpasses all other strategies across most evaluation metrics.

\begin{figure}[t]
    \centering
    \includegraphics[width=1.0\linewidth]{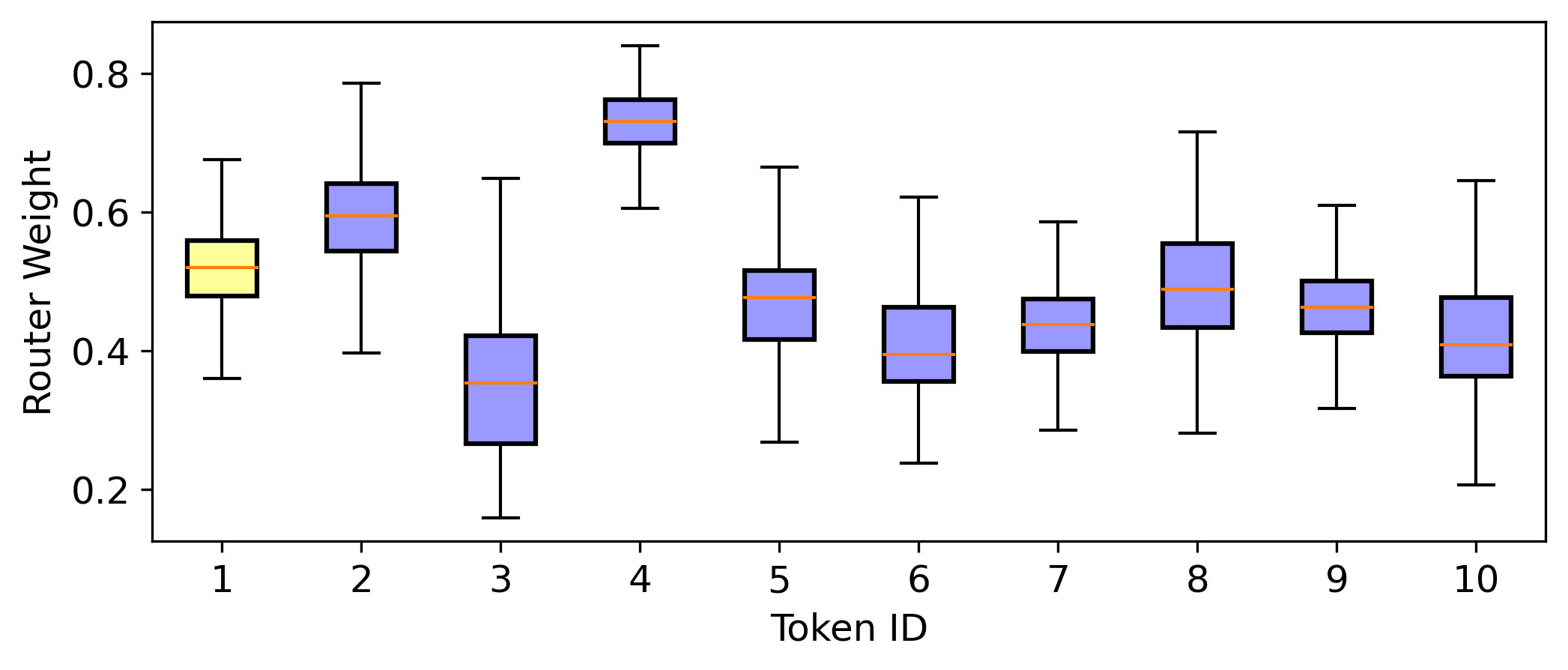}
    \vspace{-2em}
    \caption{Boxplot Visualization of token aggregation routing outcomes. We select the weights of $E_{gen}$ for analysis. The Token 1 is the CLS token, highlighted in yellow.}
    \vspace{-10pt}
    \label{fig:visualize-moe}
\end{figure}

\vspace{-15pt}
\paragraph{Visualization of Routing Outcomes.} We visualize the routing outcomes of the token aggregation router. As shown in Figure~\ref{fig:visualize-moe}, the weights assigned by the router exhibit a notable discrepancy across different token positions. Furthermore, there are subtle variations within the same token position to fit each input frame. These observations suggest that the router can adaptively adjust aggregation weights for optimal visual feature aggregation.

% \vspace{-10pt}
\subsubsection{Analysis on Token Dropping Router}
\begin{figure*}[t]
  \centering
   \includegraphics[width=1.0\linewidth]{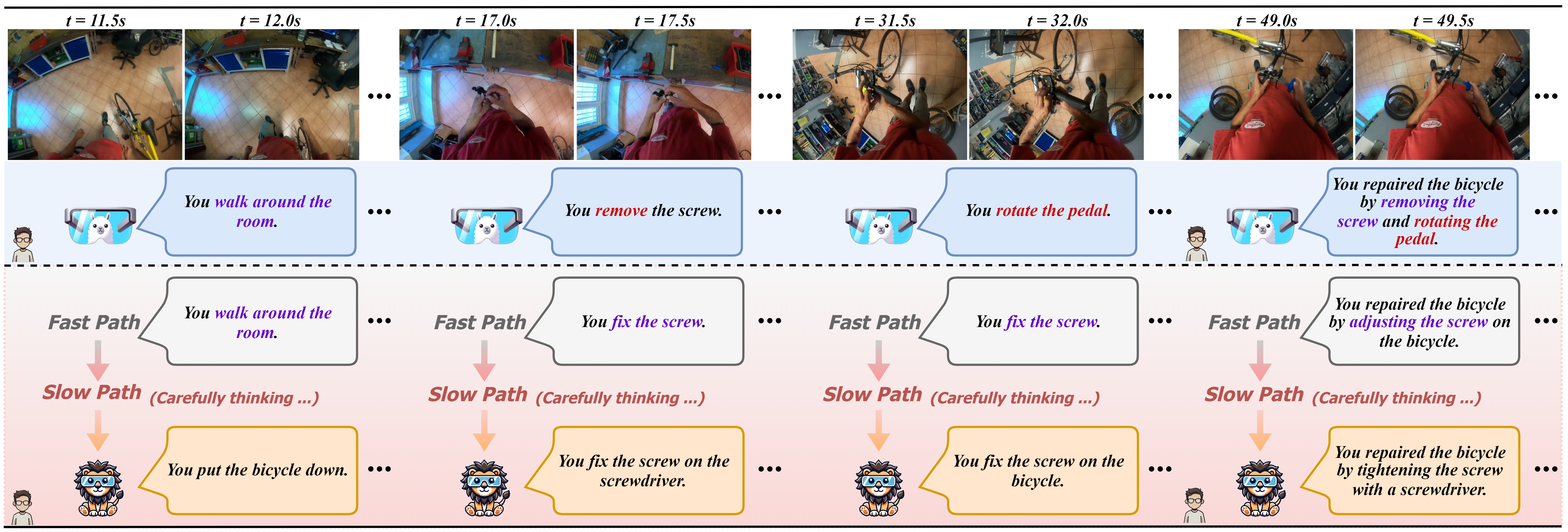}
   \vspace{-2em}
   \caption{Quantitative analysis of LIVE~\cite{VideoLLM-online}\includegraphics[height=0.9em]{fig/videollmonline.png} and LION-FS\includegraphics[height=1.5em]{fig/LION-FS.png} on the Ego4D dataset. \includegraphics[height=1.4em]{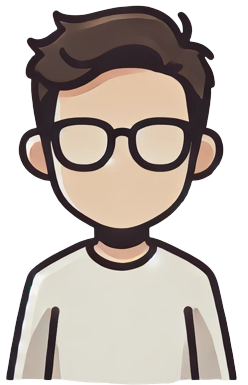} represents the questions posed by the user, such as the request ``Please narrate the video in real-time" at 0.0s and ``How did I repair the bicycle?" at 49.0s. Purple highlights indicate imprecise responses, while red highlights denote incorrect responses. LION-FS achieves progressive improvements in response precision through the integration of Fast Path and Slow Path mechanisms.}
   \vspace{-10pt}
   \label{fig:qualitative-slow}
\end{figure*}
% Table~\ref{tab:ablition_mod} presents the results of the ablation study for different settings of the redundant token drop router. We compare models with different settings against the method without token dropping (No Dropping) to find a balance between model performance and training efficiency.
% First, we investigate the impact of different placements of token dropping routers on the model. ``All Layers" and ``Interleaved Layers" indicate inserting token dropping routers in every Transformer layer and in every other layer, respectively, while ``Deep Layers" indicates inserting token dropping routers in the last two-thirds of the layers. The experimental results show that using token dropping routers allows the model to achieve limited performance loss while reducing training time. The ``Interleaved Layers'' strategy, in particular, strikes a good balance between performance and efficiency. Additionally, we discuss the impact of different token dropping rates on model performance and efficiency. From the experimental results, it is evident that even with fewer visual tokens, LION-FS achieves satisfactory results, demonstrating the critical visual selection ability of the token dropping router and highlighting the significant redundancy present in videos.

Table~\ref{tab:ablition_mod} presents the results of an ablation study on various settings of the redundant token drop router. We compare models with different configurations to the baseline method (No Dropping) in order to balance model performance and training efficiency. The random dropping method causes significant performance degradation across all metrics, highlighting the importance of a learnable dropping router. We investigate the effect of placing token dropping router at different points in the model. ``All Layers" and ``Interleaved Layers" refer to inserting token dropping router in every Transformer layer and in every other layer, respectively, while ``Deep Layers" refer to placing token dropping router in the last two-thirds of the layers. The experimental results show that using token dropping router results in minimal efficacy loss while reducing training time. Additionally, we examine the impact of varying token dropping rates on both model efficacy and efficiency. The results clearly demonstrate that even with fewer visual tokens, LION-FS achieves satisfactory efficacy, highlighting the importance of the token dropping router’s visual selection capability and revealing the significant redundancy present in video data.

\vspace{-5pt}
\subsubsection{Analysis on Multi-granularity Augmentation}

\begin{table}[t]
\caption{\textbf{Ablation Study on Multi-granularity Keyframe Augmentation.} The Baseline represents the best result of Fast-Path. ``FG tokens" refers to Fine-Grained Tokens. Experimental results show that combining Grid Tokens with Box Tokens provides the model with richer, high-quality visual information, thereby enhancing its language modeling capabilities.}
    \label{tab:ablation_slow}
    \vspace{-8pt}
    \centering
    \footnotesize
    \begin{tabular}{c|cc}
    \toprule
    % \textbf{Method} & \multicolumn{2}{c}{\textbf{Ego-Exo4D Narration Validation}} \\ & \textbf{LL-PPL} $\downarrow$ & \textbf{LM-Correctness} $\uparrow$ \\
    \textbf{Method} & \textbf{LL-PPL} $\downarrow$ & \textbf{Correctness} $\uparrow$ \\
    \midrule
    Baseline (w/o Augmentation) &2.16 &47.0\% \\
    \midrule
    + FG Tokens (1$\times$6$\times$6) &2.07 &47.3\% \\
    + Grid Tokens (4$\times$3$\times$3) &2.04 &47.8\% \\
    + Box Tokens &2.06 &47.6\% \\
    + FG Tokens (1$\times$6$\times$6) \& Box Tokens  &2.06 &47.3\% \\
    \rowcolor[HTML]{faf0e6}
    + Grid Tokens (4$\times$3$\times$3) \& Box Tokens  &\textbf{2.04} &\textbf{48.2\%} \\
    \bottomrule
    \end{tabular}
    \vspace{-18pt}
\end{table}

% To verify the effectiveness of the multi-granularity keyframe augmentation, we conducted ablation studies on the two main components: Global Uniform Augmentation and Local Adaptive Augmentation. Since Multi-Granularity Augmentation only processes keyframes of the video without involving model predictions for keyframes, it does not change the TimeDiff and Fluency metrics of the model. Therefore, this part of the exploration focuses solely on metrics related to language modeling capability and the quality of generated responses.
% To evaluate the effectiveness of Multi-Granularity keyframe augmentation, we conducted ablation studies on its two main components: Grid Tokens and Box Tokens. The goal of adding Multi-Granularity augmentation tokens is to enhance the visual information of keyframes, thereby improving the quality of the model's generated responses. In this section, we primarily focus on assessing the model's language modeling capabilities using the LL-PPL and LM-Correctness metrics. Moreover, our visual augmentation method does not affect the model’s predictions for video keyframes and therefore does not impact its efficacy in terms of TimeDiff and Fluency.

To investigate the effectiveness of Multi-Granularity keyframe augmentation, we conducted ablation studies on its two main components: Grid Tokens and Box Tokens.
We use language modeling capability metrics --- LL-PPL and LM-Correctness --- to measure the response precision improvements. As shown in Table~\ref{tab:ablation_slow}, comparing Fine-grained Tokens (1×6×6), Grid Tokens (4×3×3), and the Baseline, we observe that incorporating more comprehensive visual information into the keyframes significantly enhances the model's language modeling efficacy. Furthermore, the use of the 4×3×3 input pattern aligns the input structure of the keyframes with that of the training mode, resulting in further efficacy gains.
Furthermore, inserting Local Adaptive Augmentation multimodal prompts only at the keyframe locations provides a significant boost to LION-FS's language modeling capability. When combining Grid Tokens with Box Tokens, the Multi-Granularity Augmentation approach effectively integrates the strengths of both strategies. This combination substantially outperforms the existing model on both the LL-PPL and LM-Correctness metrics, demonstrating that Multi-Granularity Augmentation delivers richer, higher-quality visual information for the language model.
% The Multi-Granularity Augmentation, which combines Global Uniform Augmentation (4×3×3) and Local Adaptive Augmentation, effectively integrates the advantages of both approaches. It significantly surpasses the existing model on both LL-PPL and LM-Correctness metrics, indicating that Multi-Granularity Augmentation provides richer, more effective, and high-quality multimodal information for the language model.

\vspace{-10pt}
\paragraph{Qualitative Analysis} We present several examples to validate the responses' quality improvement brought by Multi-Granularity Augmentation. As shown in Figure~\ref{fig:qualitative-slow}, compared to LIVE~\cite{VideoLLM-online}, LION-FS’s Fast-Path improves the accuracy of responses, and incorporating Multi-granularity Augmentation enables the model to generate answers with richer, more fine-grained information. For example, at 49.0s, in response to the user’s question, ``How did I repair the bicycle?", LIVE incorrectly responds, ``You repaired the bicycle by removing the screw and rotating the pedal." In contrast, Fast-Path correctly reflects that the user repaired the bicycle by ``adjusting the screw." Adding Multi-granularity Augmentation further enriches the Fast-Path response, providing more detailed information.

% \begin{table}
% \caption{Ablation study slow}
%     \label{tab:ablition_moe}
%     \vspace{-8pt}
%     \centering
%     \footnotesize
%     \begin{tabular}{l|cc}
%     \toprule
%     \textbf{Method} & \multicolumn{2}{c}{\textbf{Ego-Exo4D Narration Validation}} \\ & \textbf{LL-PPL} $\downarrow$ & \textbf{LM-Correctness} $\uparrow$ \\
%     \midrule
%     Baseline(w/o Augmentation) &2.16 &47.0\% \\
%     + Fine-grained tokens (1$\times$6$\times$6) &2.07 &47.3\% \\
%     + Fine-grained tokens (4$\times$3$\times$3) &\textbf{2.04} &47.8\% \\
%     + Core tokens &2.06 &47.6\% \\
%     + Fine-grained tokens (1$\times$6$\times$6) \& Core tokens  &2.06 &47.3\% \\
%     \rowcolor[HTML]{faf0e6}
%     + Fine-grained tokens (4$\times$3$\times$3) \& Core tokens  &\textbf{2.04} &\textbf{48.2\%} \\
%     \bottomrule
%     \end{tabular}
% \end{table}

%% file: sec/5_conclusion.tex
\section{Conclusion}
\vspace{-5pt}
In this paper, we introduce LION-FS, a novel framework for online video assistant that addresses the critical challenges of both efficacy and efficiency. LION-FS employs a Fast \& Slow optimization scheme, decoupling ``intuitive" response determination from ``deliberative" response generation. To handle higher-frame-rate video streams and enhance response accuracy, the Fast Path dynamically integrates general image features with first-person video features via the Token Aggregation Router, while adaptively eliminating redundancies through the Token Dropping Router. To improve the precision of response generation, the Slow Path leverages multi-granularity augmentation of keyframes, incorporating global uniform augmentation with fine-grained pooling and local adaptive augmentation focused on areas of human-environment interaction. 

%% file: sec/X_suppl.tex
\clearpage
\setcounter{page}{1}
\maketitlesupplementary

\section*{A. Experimental Detials}
\subsection*{A.1. Architecture}
We employ two visual encoders: SigLIP and EgoVLPv2. SigLIP represents the SigLIP-large-patch16-384 model, pre-trained on the WebLi dataset at a resolution of 384×384. EgoVLPv2 is the second generation of egocentric video-language pre-training models, trained on the EgoClip version of the Ego4D dataset at a resolution of 224×224. Due to the inconsistency in token dimensions produced by these visual encoders, each set of tokens is processed through an MLP to align with the dimensionality of text tokens. The Token Aggregation Router comprises an MLP and a SoftMax layer. At the frame level, it takes the $\texttt{[VG]}$ token (i.e., the CLS token extracted by SigLIP) as input, assigning aggregation weights to all tokens derived from both visual encoders for each frame. Additionally, the Token Dropping Router employs an MLP and a SoftMax layer at the token level, assigning confidence scores to individual visual tokens. Tokens with confidence scores below a pre-defined discard threshold are deemed redundant and are subsequently dropped. For the large language model (LLM), we leverage Llama-3-8B-Instruct~\cite{llama3}, an optimized variant tailored for conversational tasks. Our online video dialogue template adheres to the instruction-tuning paradigm, extending input to encompass a multimodal fusion of interleaved visual and textual elements. In the Slow Path of LION-FS, we adopt a Faster-RCNN object detection model to detect hands and hand-interacting objects, instead of using a DETR-based model~\cite{detr}, to ensure real-time processing of video frames.

\subsection*{A.2. Data Refinement}
\begin{itemize}
\setlength{\itemsep}{0pt}
\setlength{\parsep}{0pt}
\setlength{\parskip}{0pt}

    \item \textbf{Grammar Correction.}
    For Ego-Exo4D, we reorganized and refined the annotations of short-term atomic descriptions by substituting third-person verb forms following ``C" with the second-person pronoun ``You" and the base verb form, e.g., modifying ``C stands in a house" to ``You stand in a house." Capital letters were preserved to denote other individuals (i.e., those not wearing the camera), such as in ``Lady X and man M prepare concrete in a basin," to clearly differentiate among entities. For the narrations of Ego4D, we removed markers like ``\# C" (camera wearer), ``\# O" (other individuals), and ``\# Unsure" (uncertain narration) preceding each statement, maintaining consistency with the annotation strategy applied to Ego-Exo4D.
    \vspace{5pt}
    
    \item \textbf{Dialogue Data Augmentation.} 
    We employ modified data augmentation strategies inspired by VideoLLM-online and VideoLLM-MoD: \textbf{1)} Replace a learning message with incorrect content, then correct or leave it uncorrected to train the model's ability to identify errors and respond appropriately despite misinformation. \textbf{2)} Introduce temporal inconsistencies by inserting, deleting, or replacing frames to simulate real-world frame sequence variations. \textbf{3)} Use empty strings, None, or remove messages to emulate scenarios involving missing or incomplete responses. 
    \vspace{5pt}

    \item \textbf{Dual Visual Features Alignment.}
    We employ two encoders for visual feature extraction: SigLIP extracts image tokens from frames sampled at 2 FPS, while EgoVLPv2 processes frames sampled at 8 FPS by grouping them into sets of four and extracting video tokens at 2 groups per second. To achieve temporal alignment across different frame rates, we trim the final segment of the video shorter than one second and remove any annotations exceeding the new maximum duration.
    \vspace{5pt}

\end{itemize}

\vspace{-5pt}
\subsection*{A.3. Training Settings}
All experiments are conducted using 8 × A800 80GB GPUs. We train the full MLP, Token Aggregation Router, Token Dropping Router, and LoRA~\cite{lora} embedded in each linear layer of the LLM. The batch size is set to 1 per GPU, with training conducted for 10 epochs on the Ego-Exo4D dataset and 2 epochs on the Ego4D dataset. Gradient accumulation over 32 steps is used to achieve an effectively larger batch size. We employ the AdamW~\cite{adamw} optimizer with an initial learning rate of 0.0002, using cosine learning rate scheduling with a 5\% warmup ratio. 

\begin{figure*}[t]
  \centering
   \includegraphics[width=1.0\linewidth]{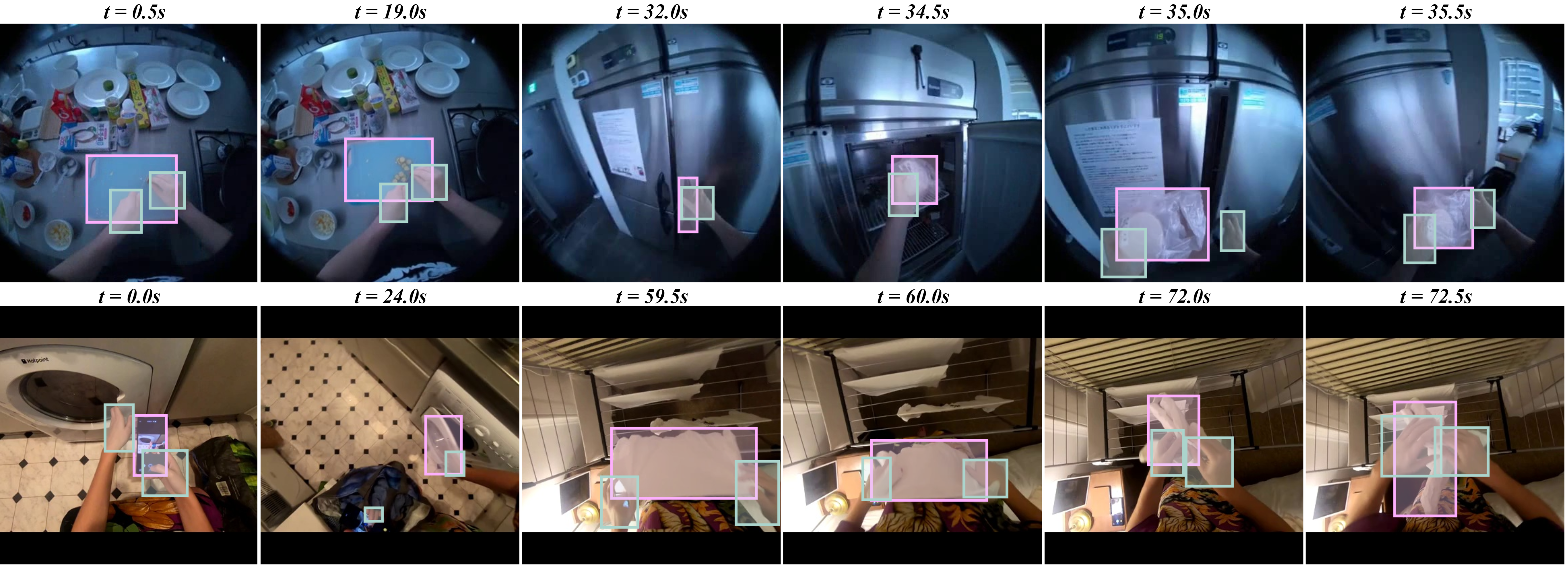}
   \vspace{-2em}
   \caption{Visualization of Local Adaptive Augmentation in the Slow Path on Ego-Exo4D and Ego4D datasets. Green bounding boxes highlight the user’s hands in the first-person view, while pink bounding boxes indicate the objects interacting with the hands. Experiments demonstrate that the object interacting with the user’s hands is often the same, and retaining only a single interaction object’s bounding box effectively reduces false detections.}
   \vspace{-10pt}
   \label{fig:sup-ho}
\end{figure*}

\subsection*{A.4. Evaluation Metrics}

\begin{itemize}
\setlength{\itemsep}{0pt}
\setlength{\parsep}{0pt}
\setlength{\parskip}{0pt}

    \item \textbf{ Language Modeling Perplexity (LM-PPL)}
    measures the quality of a model's probabilistic distribution in language modeling. It provides an indirect evaluation by assessing the probabilities assigned to each generated token. A lower LM-PPL typically indicates a stronger language modeling capability of the model.
    \vspace{5pt}
    
    \item \textbf{LM-Correctness} 
    assesses the precision of the model's language generation by focusing on the degree of alignment between generated text and reference text. By calculating the proportion of correctly generated tokens within the language sequence, it reflects the model's actual performance in generative tasks.
    \vspace{5pt}

    \item \textbf{Time Difference (TimeDiff)}
    evaluates the model’s real-time processing and temporal alignment capabilities in handling video stream inputs. It is computed as the difference between the timestamp of the response occurrence and the expected timestamp. The average TimeDiff across each dialogue turn is used as the metric.
    \vspace{5pt}

    \item \textbf{Fluency}
    assesses the integration of visual information and the naturalness and coherence of generated text. It calculates the proportion of successfully predicted tokens throughout the dialogue turns, including both response determination and response generation predictions, providing a comprehensive evaluation of language and temporal efficiency in video online dialogue.
    \vspace{5pt}

\end{itemize}

% \vspace{-5pt}
% \section*{B. Additional Quantitative Analysis}
\section*{B. Additional Experimental Analysis}

\subsection*{B.1. Additional Efficiency Evaluation}
The Fast Path of LION-FS incorporates two distinct visual encoders, enabling the aggregation of different visual tokens (using the Token Aggregation Router) and dropping redundant ones (using the Token Dropping Router), thereby enhancing the efficiency of processing high frame-rate videos. The Slow Path distinguishes keyframes (determined as responsive frames) from ordinary frames (determined as silent frames), and applies augmentation only to keyframes, minimizing efficiency impact. As shown in Table~\ref{tab:supp1}, VideoLLM-online and VideoLLM-MoD \textbf{1)} merely add tokens to every frame without optimizing specifically for keyframes, and \textbf{2)} as the input frame rate increases to 8 FPS, their inability to aggregate tokens leads to cumulative token growth, resulting in a significant rise in FLOPs. Additionally, when no frame augmentation is performed across all methods, LION-FS achieves FLOPs comparable to VideoLLM-MoD while supporting input at 8 FPS.

\begin{table}[t]
\caption{\textbf{FLOPs evaluation under the same test sample in Ego-Exo4D.} Since the code for VideoLLM-MoD has not been released, VideoLLM-MoD* is reproduced based on its paper. Both LION-FS and VideoLLM-Mod* adopt a strategy of dropping interleaved layers with a dropping ratio of $\beta = 0.5$. Leveraging the two customized routers and the keyframe augmentation strategy, LION-FS significantly reduces FLOPs even at an input frame rate of 8 FPS. Without frame augmentation, LION-FS maintains nearly constant FLOPs while supporting 8 FPS input.}
    \label{tab:supp1}
    \vspace{-5pt}
    \centering
    \small
    \setlength{\tabcolsep}{3pt}
    \begin{tabular}{c|cc|c}
    \toprule
    \textbf{Method}& \textbf{ Aug. Strategy} & \textbf{Input FPS} & \textbf{FLOPs}  \\
    \midrule
    VideoLLM-online & All frames & 2   & 55.65T  \\
    VideoLLM-online & None & 8 & 59.49T     \\
    VideoLLM-MoD* & All frames & 2  &47.00T   \\
    VideoLLM-MoD* & None & 8 &    50.29T  \\
    \rowcolor[HTML]{faf0e6}
    LION-FS & Keyframes & 8 & \textbf{ 21.53T }   \\
    \midrule
    VideoLLM-online & None & 2 & 15.45T     \\
    VideoLLM-MoD* & None & 2 & \textbf{12.28T}     \\
    \rowcolor[HTML]{faf0e6}
    LION-FS & None &8 & 12.40T     \\
    \bottomrule
    \end{tabular}
    % \vspace{-10pt}
\end{table}

\subsection*{B.2. Local Adaptive Augmentation for Box Tokens}
The primary task of an online assistant in first-person scenes is to engage in real-time dialogue with the user regarding their current actions, focusing on the interaction between the user and the environment. However, these interaction areas often constitute a small portion of the total frame in first-person videos (especially from the Ego-Exo4D \& Ego4D datasets), leading to attention dispersion in MLLMs. We address this by using bounding boxes, as shown in Figure~\ref{fig:sup-ho}, to highlight the user’s hands and their interaction with the environment, guiding the LLM to focus on these areas and improving response precision.
We first filter out the patch tokens covered by the bounding boxes in each frame, and then apply global pooling to the tokens within each bounding box to obtain a single representation per box, termed as the Box Token. We define up to three Box Tokens per frame, corresponding to the user’s two hands and an interacting object. When hands are absent or missed in detection, we replace them with a global pooling token of the frame to maintain a consistent number of tokens. In fact, when interaction regions are absent, focusing more on the global scene is often necessary.

\section*{D. Limitations}
The design of loss functions for online video dialogue modeling is hindered by the negative impact of long-tailed distributions, as the frequency of \texttt{[EOS]} occurrences significantly exceeds that of \texttt{[Assistant]} when determining response generation. Consequently, the model tends to favor predicting silence during training. To address this, we propose transforming the binary decision task into a multi-class prediction (using discrete special tokens to predict varying response probabilities) or a response probability regression task, thereby mitigating the effects of the long-tailed distribution. Additionally, we aim to introduce a fixed-length memory mechanism in LION-FS to replace the Key-Value Cache, prioritizing recent context while retaining essential historical information. This approach ensures computational efficiency, thereby enabling the handling of unlimited video lengths in online video dialogue.

\section*{E. Societal Impact and Potential Risk}
LION-FS is fine-tuned on large language models (LLMs) using the Ego-Exo4D and Ego4D datasets. Given the potential for LLMs to generate hallucinations or biased responses, and the inherent unreliability of annotations in these datasets, caution is advised when deploying LION-FS as an online video assistant. Its responses should be critically evaluated, and comprehensive safety and fairness assessments are essential before practical deployment.